\documentclass[nocompress]{spie}  

\usepackage{amsmath,amsfonts,amssymb}
\usepackage{graphicx}
\usepackage{xspace}
\usepackage{caption}
\usepackage{subcaption}
\usepackage{float}
\usepackage[table]{xcolor}
\usepackage{colortbl}
\usepackage{makecell}
\usepackage{multirow}
\usepackage[colorlinks=true, allcolors=blue]{hyperref}
\usepackage{titlesec}
\usepackage[table]{xcolor}
\usepackage{tikz}
\usepackage{tcolorbox}

\definecolor{cellred}{rgb}{1,0.6,0.6}
\definecolor{cellyellow}{rgb}{0.98, 0.89, 0.50}
\newcommand{\ours}{GS - TransUNet\@\xspace}

\title{\ours: Integrated 2D Gaussian Splatting and
Transformer UNet for Accurate Skin Lesion Analysis}
\author[a]{Anand Kumar}
\author[a]{Kavinder Roghit Kanthen}
\author[a]{Josna John}
\affil[a]{University of California, San Diego}

\authorinfo{Code: \url{https://github.com/AnandK27/GS-TransUNet}, Correspondence: ank029@ucsd.edu}

\begin{document} 
\maketitle

\begin{abstract}
We can achieve fast and consistent early skin cancer detection with recent developments in computer vision and deep learning techniques. However, the existing skin lesion segmentation and classification prediction models run independently, thus missing potential efficiencies from their integrated execution. To unify skin lesion analysis, our paper presents the Gaussian Splatting - Transformer UNet (\ours), a novel approach that synergistically combines 2D Gaussian splatting with the Transformer UNet architecture for automated skin cancer diagnosis. Our unified deep learning model efficiently delivers dual-function skin lesion classification and segmentation for clinical diagnosis. Evaluated on ISIC-2017 and PH2 datasets, our network demonstrates superior performance compared to existing state-of-the-art models across multiple metrics through 5-fold cross-validation. Our findings illustrate significant advancements in the precision of segmentation and classification. This integration sets new benchmarks in the field and highlights the potential for further research into multi-task medical image analysis methodologies, promising enhancements in automated diagnostic systems.


\end{abstract}


\section{INTRODUCTION}

Skin cancer has emerged as one of the most critical challenges in public health, with melanoma—the deadliest form—accounting for approximately 75\% of all skin cancer-related deaths~\cite{li2018dense}. Early detection and accurate diagnosis are paramount in mitigating mortality rates and improving patient outcomes. Over the past decades, medical imaging and artificial intelligence advancements have facilitated automated skin cancer diagnosis systems, demonstrating performances on par with expert dermatologists~\cite{pham2020aioutperformeddermatologistimproved, Reiter2019-tl}. However, these systems often simplify the task to binary classification, neglecting the crucial role of lesion segmentation. In practice, segmentation provides essential information about the lesion’s asymmetry, border irregularities, intensity, and size, which are indispensable for effective diagnosis.~\cite{Celebi2008-ic}

\label{sec:intro}  
\begin{figure}[hbp]
\centering
\includegraphics[width=0.5\linewidth]{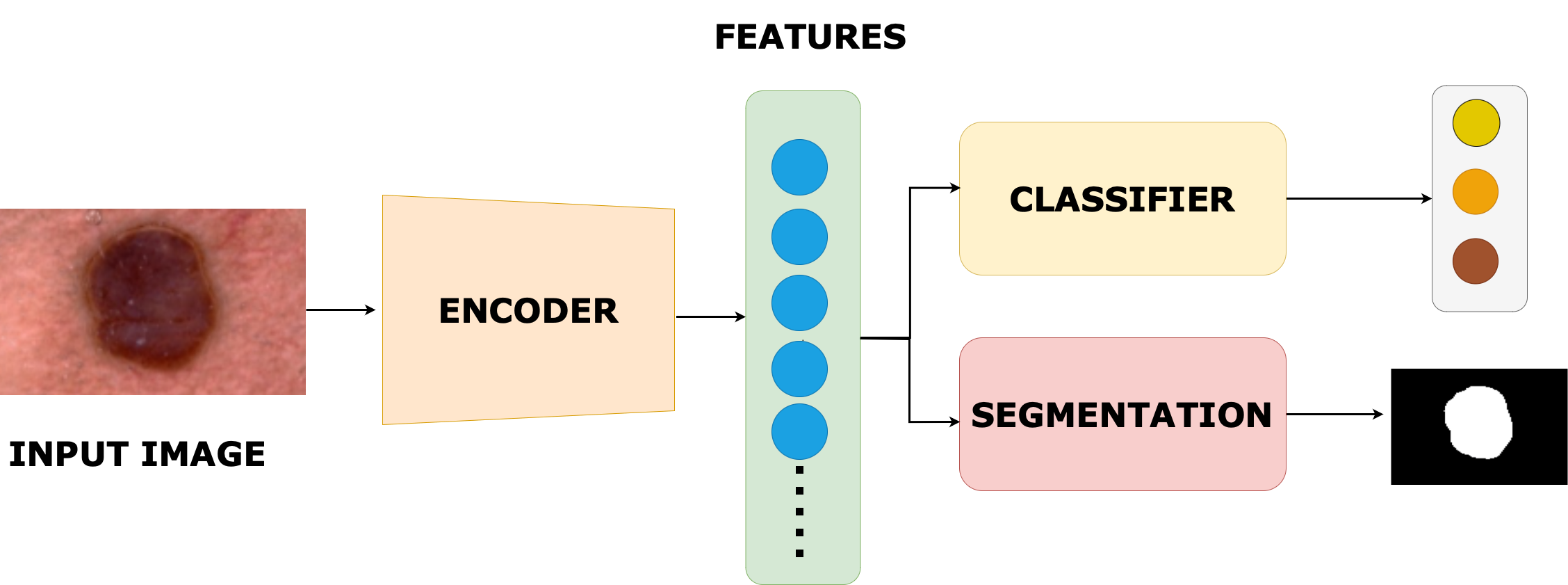} 
\vspace{1em} 
\caption{A simplified overview. The flowchart shows how our dual-task approach works for skin lesion classification and segmentation in a unified manner.}
\label{fig:model_overview}
\end{figure}
Recognizing this gap, we propose Gaussian Splatting-Transformer UNet (\ours), a novel framework designed to optimize lesion segmentation and classification tasks jointly. By integrating Convolutional Neural Networks (CNNs) and Vision Transformers (ViT)~\cite{vit}, \ours captures both local details (lesion texture and color) and global context (lesion shape and spatial relationships), as shown in Fig.\ref{fig:model_overview}. Unlike previous models~\cite{mt_trans}, \ours introduces 2D Gaussian Splatting, a fully differentiable method for generating precise binary masks of elliptical lesions, improving segmentation accuracy even in cases without ground truth masks. This integration enhances segmentation consistency and ensures robust classification performance by focusing on the lesion region and ignoring artifacts such as hair, scales, and tapes.

Building upon the foundational architecture of UNet~\cite{unet}, \ours addresses UNet’s inherent limitation of modeling long-range dependencies by incorporating Transformer layers into the encoder. These layers leverage Multi-Head Self-Attention (MSA) to establish correlations over long distances, enabling the segmentation of expansive or irregular regions commonly observed in dermoscopic images. The framework also introduces innovative loss strategies, including Dual Task Consistency (DTC),~\cite{dtc} to ensure alignment between segmentation and classification outputs, further strengthening the network’s robustness. We validate \ours on the ISIC-2017~\cite{isic} and PH2~\cite{ph2} datasets, demonstrating its efficacy in both segmentation and classification. Our results highlight significant improvements in metrics such as the DICE coefficient, Jaccard index, precision, recall, and classification accuracy, surpassing state-of-the-art models. Moreover, the framework’s computational efficiency and consistency losses ensure practical applicability for real-world dermatological diagnosis.

To summarize, our framework distinguishes itself from existing approaches through these advancements:

\begin{enumerate}
    \item \ours introduces a dual-task model that jointly optimizes segmentation and classification tasks. By leveraging Gaussian splatting, the model seamlessly integrates these two tasks, enabling mutual reinforcement and improving overall performance.

    \item We introduce a novel approach that generates segmentation masks through two parallel networks using Gaussian splatting and signed distance fields. This dual-path design ensures that the generated masks are consistent and robust to noise, addressing common challenges in medical image segmentation.

    \item \ours achieved a 2.5\% improvement in accuracy, setting new benchmarks in classification and segmentation tasks.

\end{enumerate}

\subsection{Related Works}
Skin cancer analysis has had various advancements over the years, which have been influential in designing our model's architecture. We highlight and compare other approaches with ours in the following section.

\subsubsection{Skin Lesion Segmentation}
The evolution from thresholding and active contour models ~\cite{hemalatha2018active, ravichandran2009color,yogarajah2010dynamic} to deep neural networks has revolutionized skin lesion segmentation, as seen in the 19-layer fully convolutional network employing Jaccard distance loss~\cite{nasr2019dense} and star shape priors for global structure preservation~\cite{mirikharaji2018star}. With inspiration from~\cite{centernet}, \ours advances segmentation by introducing 2D Gaussian splatting and dual-task learning, ensuring robust and consistent mask generation without reliance on specific priors or loss functions.

\subsubsection{Skin Image Classification}
The advent of deep neural networks has significantly enhanced skin image classification, transitioning from manual feature engineering~\cite{hagerty2019deep} to automated methods like attention residual learning~\cite{zhang2018skin}. \ours employs Vision Transformers (ViT) to enhance classification by capturing long-range dependencies and global context in skin images.

\subsubsection{Multi-task Learning}
Initial multi-task learning approaches like MB-DCNN~\cite{mb_dcnn} and MT-TransUNet~\cite{yu2016automated} showcased the benefits of integrating segmentation and classification tasks but often relied on separate or token-based methods. \ours advances multi-task learning with an end-to-end framework that enhances task interdependence through dual-task consistency and robust mask generation.

\subsubsection{Advanced Vision Transformers in Dermatology}
Vision Transformers (ViTs)~\cite{vit}, originally developed for natural language processing, have redefined the analysis of dermatological images by capturing long-range dependencies and global context. \ours builds on ViTs within the Transformer UNet architecture, leveraging their strengths for dermatological image analysis.

\subsubsection{Task Regularization for Enhanced Learning}
Consistency regularization has proven effective in semi-supervised learning contexts, as demonstrated by MT-TransUNet~\cite{dtc,robust}, which aligns segmentation and classification through dual-task and attended region consistency losses. \ours enhances this strategy with 2D Gaussian splatting and dual-task consistency loss, achieving robust performance without heavy dependence on large labeled datasets.

The development of \ours model stands as a testament to the confluence of deep learning innovations, transformer architectures, and consistency regularization strategies in the domain of skin lesion analysis and pushing the boundaries for obtaining robust and reliable results.



\section{METHODS}

\subsection{Model Architecture}
\begin{figure}[htbp]
\centering
\includegraphics[width=\linewidth]{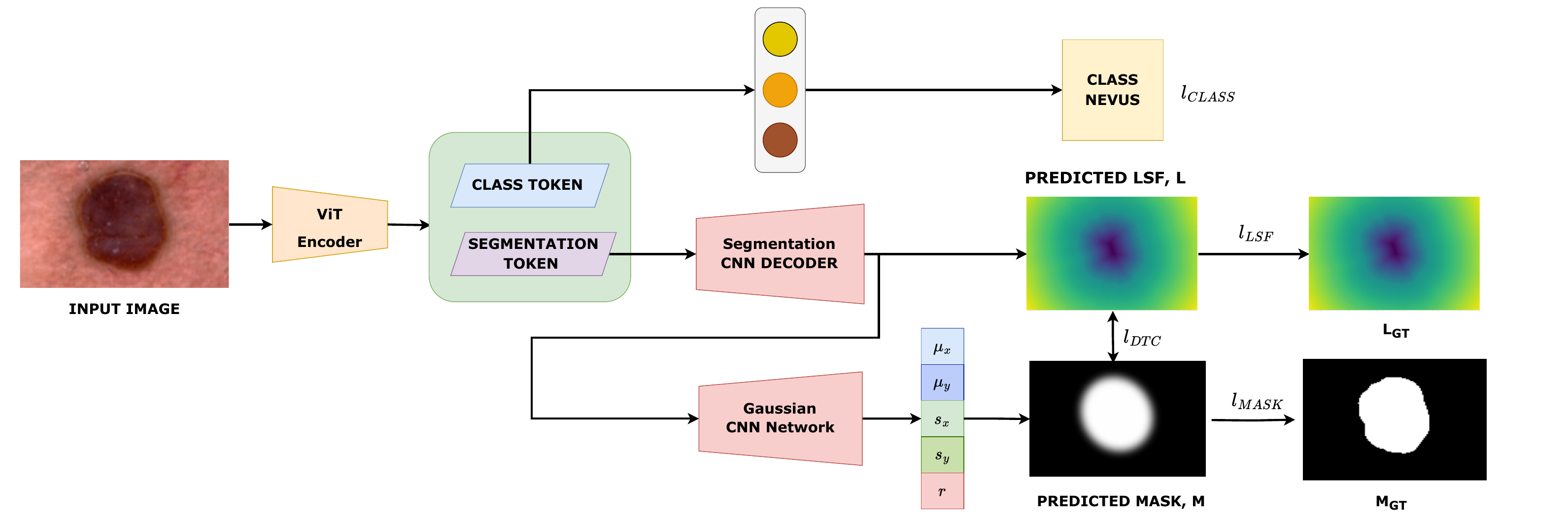} 
\vspace{1em} 
\caption{Architecture of the \ours model, combining Vision Transformer with UNet for enhanced segmentation and classification.The input image is passed through a pre-trained vision transformer encoder to obtain the class(global) and segment(local) tokens. These tokens are used for simultaneous classification and segmentation. The classification is done using an MLP, and the segmentation mask is predicted using two parallel networks: (i) CNN Decoder to obtain level set function and (ii) CNN Gaussian Network to obtain binary mask. More explanation in Appendix \ref{sec:app_enc_dec}. }
\label{fig:model_architecture}
\end{figure}

The \ours model as shown in Fig.~\ref{fig:model_architecture}combines the strengths of Convolutional Neural Networks (CNNs) and Vision Transformers to capture local and global features from skin lesion images. The architecture integrates a Transformer UNet structure, which effectively models long-range dependencies, a key requirement for handling the complex patterns found in skin lesions. This is achieved by employing the Vision Transformer as the encoder and utilizing UNet-style skip connections to preserve spatial information during upsampling.

The input image \( I \) is first downsampled by a factor of 4 using a ResNet50 backbone ~\cite{resnet} to create a feature map \( F \) of size \( H' \times W' \times C' \). These features are split into patches and passed through the Vision Transformer layers~\cite{vit} to generate embedding tokens, which are then used to predict class labels and generate segmentation masks. 

The input features are converted into patches of size \( P \times P \) and then mapped to a sequence of linear embeddings. These embeddings form the input tokens for the Transformer:
\begin{align*}
\mathbf{X} &= \text{Reshape}\left(\text{Conv}\left(I\right)\right) \\
\mathbf{Z} &= \mathbf{A}\mathbf{X}
\end{align*}
where \(\mathbf{A}\) is a learnable linear map that projects the input patch into a \( D \)-dimensional embedding space.

The Transformer layer primarily comprises Multihead Self-Attention (MSA)~\cite{attention} and a Feed-Forward Network (FFN), with Layer Normalization (LN) applied after each operation with detailed architecture given in Appendix~\ref{sec:app_enc_dec}. The outputs of each layer are given by:
\begin{align*}
\mathbf{Z_i} &= \text{MSA}(\text{LN}(\mathbf{Z_{i-1}})) + \mathbf{Z_{i-1}} \\
\mathbf{Z_i} &= \text{FFN}(\text{LN}(\mathbf{Z_i})) + \mathbf{Z_i}
\end{align*}
The tokens obtained after the transformer layers are split into two: the \textit{classification}(global) token, which is the first index of the tokens \((\mathbf{Z_i}[0, :])\) and \textit{segmentation}(local) tokens, the remaining tokens \((\mathbf{Z_i}[1:, :])\).

\subsection{2D Gaussian Splatting}
\begin{figure}[htbp]
    \centering
    \includegraphics[width = 0.6\columnwidth]{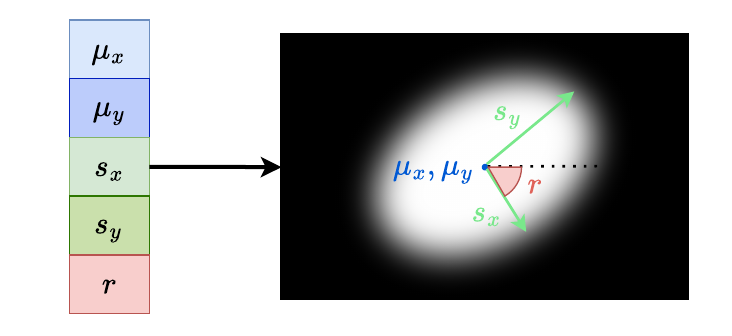}
    \vspace{1em} 
    \caption{Illustrative diagram for generating binary masks using 2D Gaussian Splatting. The Gaussian Splats are generated using $6$ features: $\mu_x, \mu_y, s_x, s_y, r$}
    \label{fig:gs}
\end{figure}

A novel 2D Gaussian Splatting method as shown in Fig.~\ref{fig:gs} is used to generate segmentation masks, focusing on creating consistent and accurate representations of elliptical lesion boundaries. This method utilizes Gaussian functions to model the shape and scale of lesions, improving the precision of boundary delineation compared to traditional deconvolution techniques. This splatting method is particularly effective in highlighting the asymmetry and irregular borders characteristic of melanoma.

To model the lesions, we use a Gaussian function for a pixel($x,y$) in the image,\\\( G(x, y) = \exp \left(-\frac{1}{2} \begin{bmatrix} x - \mu_x \\ y - \mu_y \end{bmatrix}^T \Sigma^{-1} \begin{bmatrix} x - \mu_x \\ y - \mu_y \end{bmatrix} \right) \), where \(\Sigma = R \begin{bmatrix} s_x^2 & 0 \\ 0 & s_y^2 \end{bmatrix} R^T\) and \( R = \begin{bmatrix} \cos r & -\sin r \\ \sin r & \cos r \end{bmatrix} \) is the rotation matrix for angle \( r \). The input consists of $6$ features: center coordinates ($\mu_x$ and $\mu_y$), size of the splat ($s_x$ and $s_y$) and rotation ($r$) in radians.

This Gaussian is ``splatted" onto the image grid to form the segmentation mask \( M \). The elements of the covariance matrix are adjusted based on the scale and rotation parameters derived from the network.
\subsection{Loss Functions and Training Strategy}
Our model employs several loss functions to enhance segmentation and classification accuracy. Focal Loss~\cite{focal} addresses class imbalance by emphasizing difficult samples, defined as:
\[
\text{Focal Loss}(p,y) = -\sum_{i=1}^{K} \left( y_i \log(p_i) (1 - p_i)^\gamma \alpha_i + (1 - y_i)\log(1 - p_i)p_i^\gamma (1 - \alpha_i) \right),
\]

where \(\alpha\) and \(\gamma\) are hyperparameters, leading to the classification loss \(l_{\text{CLASS}} = \text{FocalLoss}(\hat{z}, z_{GT})\). Dice Loss~\cite{dice} enhances segmentation by maximizing the overlap between predicted and true masks:
\[
\text{Dice Loss}(X, Y) = 1 - \frac{2|X \cap Y|}{|X| + |Y|}.
\]

$L_2$ Loss aligns predicted masks with ground truth when available:
\[
l_{MASK} = L_2(M, M_{GT}), \quad l_{LSF} = L_2(L, L_{GT}).
\]
The level set function (\(L\)) encourages dual task consistency by representing distances from the boundary. It is derived from the binary mask using:
\[
T(x) = \begin{cases} 
-\inf_{y \in \partial S} ||x - y||_2, & x \in S_{\text{in} } \\
0, & x \in \partial S \\
+\inf_{y \in \partial S} ||x - y||_2, & x \in S_{\text{out} }
\end{cases} = \text{sign}(x) \cdot \inf_{y \in \partial S} ||x - y||_2
\]


Each pixel in \(L\) is valued by its minimum distance to the boundary \(\partial S\), with directionality indicating whether it is inside (\(-\)) or outside (\(+\)) the boundary. 

When segmentation ground truth is available, \(L\) and the binary mask \(M\) are aligned with the ground truth using L2 loss:
\[
l_{MASK} = L_2(M, M_{GT}), \quad l_{LSF} = L_2(L, L_{GT})
\]
The Dual-Task Consistency Loss ensures consistency between binary masks and level set functions by converting the level set function (\(L\)) back to a binary mask (\(M'\)) and comparing it with the generated binary mask (\(M\)):
\[
M' = \text{Sigmoid}(k \cdot L), \quad l_{DTC} = L_2(M', M),
\]
where \(k\) is typically set to 1500.

Our overall loss function combines these elements, optimizing segmentation and classification with the equation:
\[
l_{\text{TOTAL}} = l_{\text{CLASS}} + \lambda_M \cdot l_{MASK} + \lambda_L \cdot l_{LSF} + \lambda_{DTC} \cdot l_{DTC} + \lambda_{Dice} \cdot \text{Dice Loss}(M, M_{GT}),
\]

where \(\lambda_M\), \(\lambda_L\), \(\lambda_{DTC}\), and \(\lambda_{Dice}\) are weighting factors balancing each loss term's contribution. The training strategy uses a batch size of $8$ with data augmentation (flipping, rotation, noise) to improve generalization. These loss functions guide the model to learn effectively from both labeled and unlabeled data, enhancing robustness and accuracy.

\section{EXPERIMENTS}

\subsection{Datasets and Evaluation Metrics}
The model was evaluated on the ISIC-2017 and PH2 datasets, which provide a comprehensive set of dermoscopic images with corresponding segmentation masks and classification labels. The evaluation metrics include the Jaccard index (JA), Dice coefficient (DI), pixel-wise accuracy (pixel-AC), sensitivity (pixel-SE), specificity (pixel-SP), and classification accuracy (AC).

\subsubsection{ISIC-2017 Dataset}
The ISIC-2017 dataset contains 2,750 dermoscopic images with annotations for training and testing lesion segmentation and classification algorithms~\cite{isic}. This dataset presents a variety of lesion types, providing a robust foundation for model validation.

\subsubsection{PH2 Dataset}
The PH2 dataset consists of 200 dermoscopic images with detailed segmentation masks and classification labels~\cite{ph2}. It includes images of melanomas, common nevi, and atypical nevi, offering a diverse sample for evaluating model performance.


\subsection{Implementation Details}\label{sec:imp}

We use a batch size of $8$, split into two mini-batches of size $4$, trained on $l_{SEG}$ and $l_{NON-SEG}$ respectively, enhancing robustness to samples without segmentation masks. Networks are pre-trained on ImageNet1K~\cite{imagenet}. Weights are updated with the Adam optimizer~\cite{adam} using default momentum and weight decay. Models are trained for 80 epochs with an initial learning rate of $10^{-5}$, which decreases on plateau. The transformer-based networks use $16\times16$ image patches with ResNet50~\cite{resnet} as the backbone and 4 transformer layers. We set $\lambda_M = 0.25$, $\lambda_L = 0.5$, $\lambda_{Dice} = 0.5$, and exponentially increase $\lambda_{DTC}$ over epochs. Focal loss parameters $\gamma$ and $\alpha$ are set to $2$ and $0.25$.

Model performance is evaluated using 5-fold cross-validation, where the dataset is split into $5$ equal parts, and the model is trained from scratch $5$ different times by keeping one of the parts as the testset and the rest $4$ as trainset. During training, images are augmented with vertical and horizontal flips, resizing, center-cropping, rotation, and Gaussian noise, each with a 0.5 probability. Input images are sized $224\times224$. Each image undergoes three rounds of augmentation for testing, and results are averaged.

Our models are implemented using PyTorch and Sklearn, trained with CUDA on an RTX 3090 GPU with 24 GB RAM.
\renewcommand{\arraystretch}{1.5}
\begin{table*}[t]
\centering
\resizebox{\textwidth}{!}{ 
\begin{tabular}{l|c|c|c|c|c|c|c|c|c|c|c}
\hline
\multirow{2}{*}{\textbf{Models}} & \multicolumn{5}{c|} {\textbf{Melanoma Classification}} & \multicolumn{5}{c|} {\textbf{Keratosis Classification}} & \textbf{Average} \\
\cline{2-11}
 & \textbf{M\_AC} & \textbf{M\_AUC} & \textbf{M\_AP} & \textbf{M\_SE} & \textbf{M\_SP} & \textbf{K\_AC} & \textbf{K\_AUC} & \textbf{K\_AP} & \textbf{K\_SE} & \textbf{K\_SP} & \textbf{Accuracy} \\
\Xhline{1.5pt}
\Xhline{1.5pt}
RegNet~\cite{xu2021regnet} & 0.860 & 0.812 & 0.596 & 0.400 & \cellcolor{cellred}0.975 & \cellcolor{cellyellow}0.940 & \cellcolor{cellred}0.987 & \cellcolor{cellred}0.957 & \cellcolor{cellred}0.904 & 0.953 & 0.806 \\
\hline
EffNet~\cite{effnet} & 0.860 & 0.861 & 0.740 & 0.400 & \cellcolor{cellred}0.975 & 0.926 & \cellcolor{cellyellow}0.979 & 0.946 & 0.833 & 0.962 & 0.813 \\
\hline
ViT~\cite{vit} & \cellcolor{cellred}0.886 & \cellcolor{cellyellow}0.898 & \cellcolor{cellyellow}0.752 & \cellcolor{cellred}0.600 & 0.958 & \cellcolor{cellred}0.960 & 0.968 & 0.943 & 0.714 & \cellcolor{cellred}0.972 & 0.813 \\
\Xhline{1.5pt}
SDL~\cite{zhang2018skin} & 0.876 & - & - & - & - & 0.933 & - & - & - & - & 0.814 \\
\hline
MB-DCNN~\cite{mb_dcnn} & 0.856 & 0.892 & 0.703 & 0.443 & 0.963 & 0.914 & 0.931 & 0.884 & 0.872 & 0.823 & 0.807 \\
\hline
MT-TransUNet~\cite{mt_trans} & 0.873 & 0.895 & 0.678 & 0.466 & \cellcolor{cellred}0.975 & 0.926 & 0.959 & 0.901 & \cellcolor{cellyellow}0.881 & 0.944 & \cellcolor{cellyellow}0.826 \\
\hline
\ours w/o Focal Loss(Ours) & \cellcolor{cellyellow}0.880 & \cellcolor{cellred}0.905 & 0.749 & \cellcolor{cellred}0.600 & 0.950 & 0.926 & 0.954 & 0.939 & 0.761 & \cellcolor{cellyellow}0.991 & \cellcolor{cellyellow}0.826 \\
\hline
\ours w Focal Loss(Ours) & \cellcolor{cellred}0.886 & 0.889 & \cellcolor{cellred}0.756 & \cellcolor{cellyellow}0.567 & \cellcolor{cellyellow}0.966 & \cellcolor{cellyellow}0.940 & 0.969 & \cellcolor{cellyellow}0.953 & 0.880 & 0.969 & \cellcolor{cellred}0.846 \\
\hline
\end{tabular}
} 
\center\caption{Melanoma and Keratosis Classification Results for different networks with the best highlighted using red and the second best using yellow for each metric baseline networks. AC, AUC, AP, SE, and SP refer to Accuracy, Area Under Curve, Average Precision, Sensitivity, and Specificity respectively.}
\label{table:1}
\end{table*}

\subsection{Performance Analysis}\label{sec:methods}
The \ours model demonstrates substantial improvements over state-of-the-art and baseline methods. For baseline methods, we used EffNet (Efficient Net)~\cite{effnet}, RegNet (Regulated Net)~\cite{xu2021regnet}, and ViT (Vision Transformer)~\cite{vit} models pre-trained on ImageNet~\cite{imagenet}. We also performed classification using traditional statistical models such as Support Vector Machines (SVMs)~\cite{svm}, AdaBoost~\cite{adaboost}, Extreme Gradient Boosting (XGBoost)~\cite{xgb}, etc. and their results are given in Appendix~\ref{sec:app_stat}. The state-of-the-art methods include MT-TransUNet (Multi-Task Transformer UNet)~\cite{mt_trans}, MB-DCNN (Mutual Bootstrapping Deep Convolutional Neural Networks)~\cite{mb_dcnn} and SDL (Synergic Deep Learning)~\cite{zhang2018skin}. For melanoma classification, the model achieved an accuracy of $88.6\%$ with an AUC of $0.889$, while keratosis classification reached an accuracy of $94.0\%$ with an AUC of $0.969$ as shown in Table~\ref{table:1}. These results highlight the model's efficacy in distinguishing between skin lesions, a crucial capability for accurate diagnosis.

Table \ref{table:1} provides a detailed comparison of classification metrics across different models, illustrating \ours's superior performance. The model's segmentation results, presented in Table \ref{table:2}, further emphasize its ability to generate precise and consistent lesion boundaries, crucial for reliable diagnosis where our model outperforms other state-of-the-art methods such as MB-DCNN and MT-TransUNet on pixel-wise accuracy and pixel-wise specificity resulting fewer false positives.

\begin{figure}[htbp]
\centering
\begin{subfigure}[b]{0.48\linewidth} 
\includegraphics[width=\linewidth]{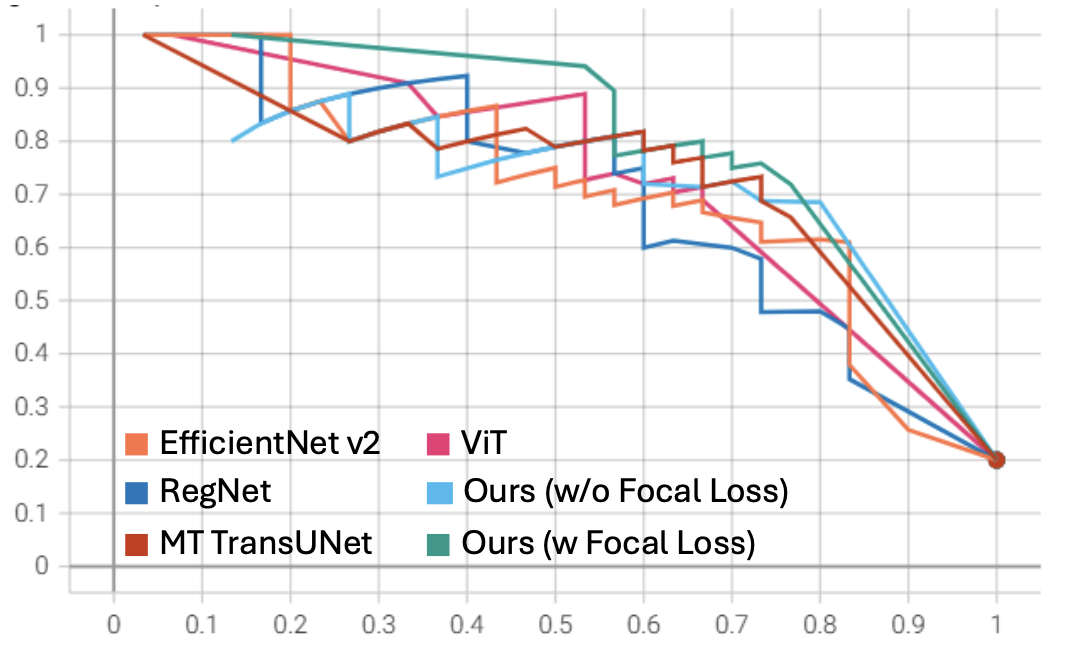}
\caption{Melanoma}
\label{fig:R12}
\end{subfigure}
\hfill
\begin{subfigure}[b]{0.48\linewidth} 
\includegraphics[width=\linewidth]{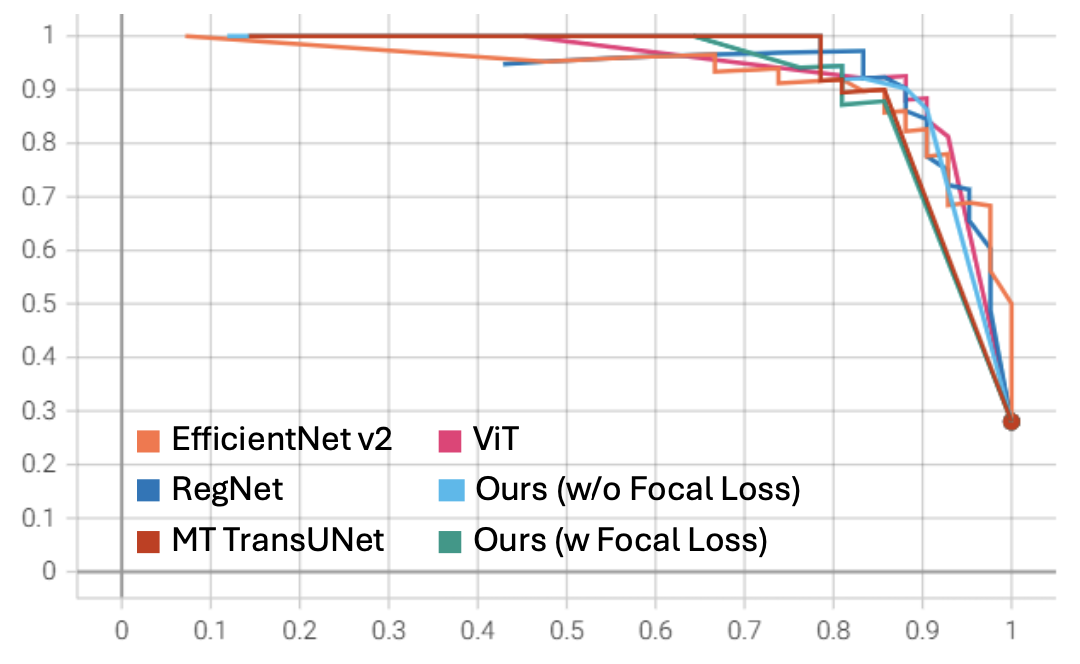}
\caption{Keratosis}
\label{fig:R22}
\end{subfigure}
\vspace{1em} 
\caption{Precision-Recall Curves for Melanoma and Keratosis cases}
\label{fig:prraphs}
\end{figure}

The Precision-Recall (PR) curves for both melanoma Fig.~\ref{fig:R12} and keratosis classification Fig.~\ref{fig:R22} were instrumental in visualizing the model's efficacy. Our model with focal loss sustained higher precision across an extended range of recall values than its counterpart without focal loss and other models. These curves emphasized the model's capability to maintain a high true positive rate with minimal false positives, particularly beneficial in medical contexts where the stakes of misdiagnosis are high.

\begin{table*}
\begin{center}
\resizebox{\textwidth}{!}{
\begin{tabular}{l *{5}{|>{\centering\arraybackslash}m{25mm}} }
\hline
\multirow{2}{*}{\textbf {Methods}} & \textbf {Jaccard}  & \textbf {DICE} & \multirow{2}{*}{\textbf {Pixel-AC}} & \multirow{2}{*}{\textbf {Pixel-SE}} & \multirow{2}{*}{\textbf {Pixel-SP}} \\
& \textbf{(JA)}& \textbf {Score}& & & \\
\Xhline{1.5pt}
MB-DCNN~\cite{mb_dcnn}& 0.595 & 0.730 & 0.914 & 0.839 & 0.921 \\
\hline
MT-TransUNet~\cite{mt_trans} & \cellcolor{cellred}0.606 & \cellcolor{cellred}0.733 & 0.928 & \cellcolor{cellred}0.797 & \cellcolor{cellyellow}0.930 \\
\hline
\ours w/o Focal Loss(Ours) & \cellcolor{cellyellow}0.580 & \cellcolor{cellyellow}0.712 & \cellcolor{cellyellow}0.932 & \cellcolor{cellyellow}0.692 & \cellcolor{cellred}0.973 \\
\hline
\ours w Focal Loss(Ours) & 0.554 & 0.689 & \cellcolor{cellred}0.934 & 0.630 & \cellcolor{cellred}0.973 \\
\hline
\end{tabular}
}
\end{center}
\vspace{1em} 
\caption{Segmentation Results for different networks with the best highlighted using red and the second best using yellow for each metric.}
\label{table:2}
\end{table*}

\begin{figure*}[ht]
\centering
\begin{subfigure}{0.16\linewidth}
\centering
\includegraphics[width=\linewidth]{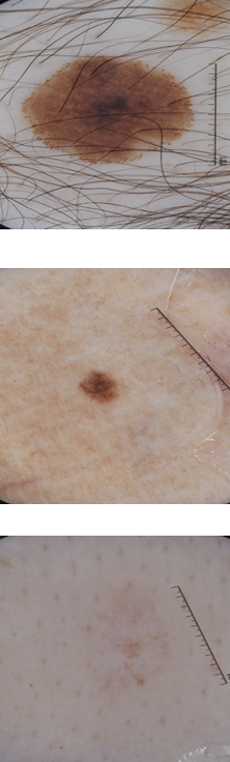}
\caption{Skin Image}
\end{subfigure}
\begin{subfigure}{0.16\linewidth}
\centering
\includegraphics[width=1.025\linewidth]{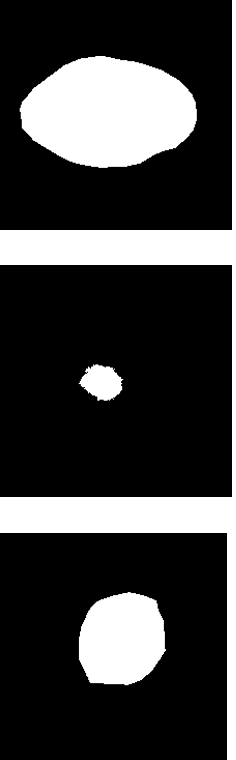}
\caption{Ground Truth}
\end{subfigure}
\begin{subfigure}{0.16\linewidth}
\centering
\includegraphics[width=\linewidth]{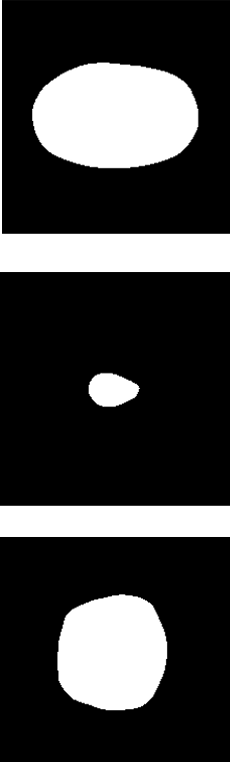}
\caption{MT-TransUNet}
\end{subfigure}
\begin{subfigure}{0.16\linewidth}
\centering
\includegraphics[width=\linewidth]{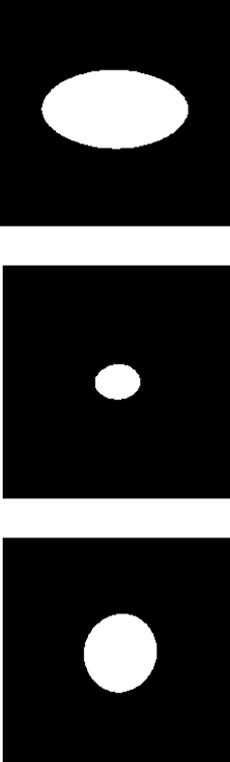}
\caption{Ours}
\end{subfigure}
\vspace{1em} 
\caption{Qualitative results of our network (\ours) and MT-TransUNet with each row showing a different test case.}
\label{fig:Results_cases}
\end{figure*}

\subsection{Output Visualization}
\begin{figure*}[ht]
\centering
\begin{subfigure}{0.16\linewidth}
  \centering
  \includegraphics[width=\linewidth]{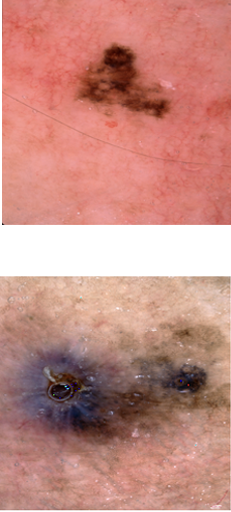}
  \caption{Skin Image}
\end{subfigure}
\hspace{2pt}
\begin{subfigure}{0.16\linewidth}
  \centering
  \includegraphics[width=\linewidth]{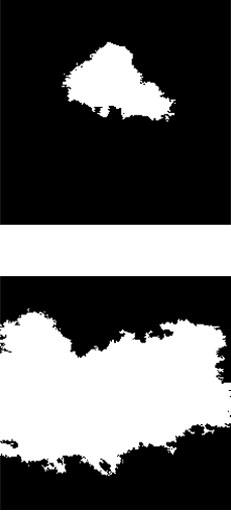}
  \caption{Ground Truth}
\end{subfigure}
\hspace{2pt}
\begin{subfigure}{0.16\linewidth}
  \centering
  \includegraphics[width=\linewidth]{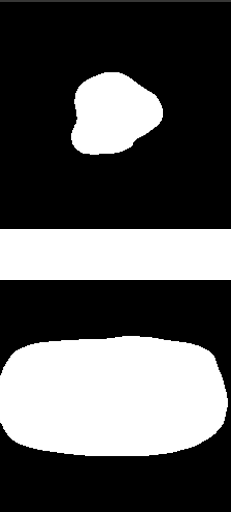}
  \caption{MT-TransUNet}
\end{subfigure}
\hspace{2pt}
\begin{subfigure}{0.16\linewidth}
  \centering
  \includegraphics[width=\linewidth]{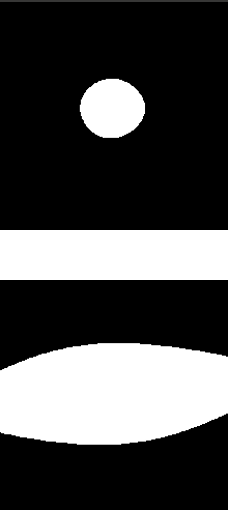}
  \caption{Ours}
\end{subfigure}
\vspace{1em} 
\caption{Failure Cases of our network (\ours) and MT-TransUNet for different test cases.}
\label{fig:failurecases}
\end{figure*}

Figs.~\ref{fig:Results_cases} and \ref{fig:failurecases} display qualitative results from the \ours model. The segmentation outputs clearly delineate lesion boundaries, with our model ensuring robust handling of artifacts like hair and shadows, improving both segmentation and classification outputs. However, challenges remain in handling irregularly shaped lesions, as the Gaussian splatting method inherently favors elliptical structures, leading to oversimplified masks. These failure cases illustrate areas for further improvement, such as handling complex lesion shapes or varying lighting conditions. Despite these limitations, \ours demonstrates superior performance and generalization, with future improvements needed to address these edge cases.

\section{DISCUSSION AND CONCLUSION}

In this paper, we present \ours, a unified approach for skin cancer classification and segmentation by utilising ViT and Gaussian Splats. \ours challenges the norm by implementing a parallel mask rendering technique that uses the probabilistic properties of 2D Gaussian splatting to enhance segmentation mask prediction. The model's architecture leverages the long-range dependencies handled by transformers and the detailed localization afforded by Gaussian splatting, allowing for refined interpretation of dermoscopic images. Through this dual-task method, \ours model represents a significant advancement in automated dermatological diagnosis. The model achieves higher accuracy and computational efficiency by integrating segmentation and classification tasks than traditional methods. This innovative approach sets new benchmarks in the field and underscores the potential of integrating deep learning techniques for improved patient outcomes in skin cancer detection.

Since Gaussian Splats generate solely elliptical masks, our technique will fail when the ground truth mask is not spherical. However, that is not true for skin cancer segmentation, as most skin blobs are circular. While our current work focuses on developing a dual-task approach to skin cancer analysis, future work can focus on generalizing our approach for other scenarios and potentially integrating 3D Gaussian Splats in MRI segmentation.



\bibliography{report} 

\begin{thebibliography}{10}

\bibitem{li2018dense}
Li, H., He, X., Zhou, F., Yu, Z., Ni, D., Chen, S., Wang, T., and Lei, B., ``Dense deconvolutional network for skin lesion segmentation,'' {\em IEEE journal of biomedical and health informatics}~{\bf 23}(2),  527--537 (2018).

\bibitem{pham2020aioutperformeddermatologistimproved}
Pham, C.~T., Luong, M.~C., Hoang, D.~V., and Doucet, A., ``Ai outperformed every dermatologist: Improved dermoscopic melanoma diagnosis through customizing batch logic and loss function in an optimized deep cnn architecture,'' (2020).

\bibitem{Reiter2019-tl}
Reiter, O., Rotemberg, V., Kose, K., and Halpern, A.~C., ``Artificial intelligence in skin cancer,'' {\em Curr. Dermatol. Rep.}~{\bf 8},  133--140 (Sept. 2019).

\bibitem{Celebi2008-ic}
Celebi, M.~E., Kingravi, H.~A., Iyatomi, H., Aslandogan, Y.~A., Stoecker, W.~V., Moss, R.~H., Malters, J.~M., Grichnik, J.~M., Marghoob, A.~A., Rabinovitz, H.~S., and Menzies, S.~W., ``Border detection in dermoscopy images using statistical region merging,'' {\em Skin Res. Technol.}~{\bf 14},  347--353 (Aug. 2008).

\bibitem{vit}
Dosovitskiy, A., Beyer, L., Kolesnikov, A., Weissenborn, D., Zhai, X., Unterthiner, T., Dehghani, M., Minderer, M., Heigold, G., Gelly, S., et~al., ``An image is worth 16x16 words: Transformers for image recognition at scale,'' {\em arXiv preprint arXiv:2010.11929}  (2020).

\bibitem{mt_trans}
Chen, J., Chen, J., Zhou, Z., Li, B., Yuille, A., and Lu, Y., ``Mt-transunet: Mediating multi-task tokens in transformers for skin lesion segmentation and classification,'' (2021).

\bibitem{unet}
Ronneberger, O., Fischer, P., and Brox, T., ``U-net: Convolutional networks for biomedical image segmentation,'' in [{\em Medical image computing and computer-assisted intervention--MICCAI 2015: 18th international conference, Munich, Germany, October 5-9, 2015, proceedings, part III 18}{\nolinebreak\hspace{0.1em}]},   234--241, Springer (2015).

\bibitem{dtc}
Luo, X., Chen, J., Song, T., and Wang, G., ``Semi-supervised medical image segmentation through dual-task consistency,'' in [{\em Proceedings of the AAAI conference on artificial intelligence}{\nolinebreak\hspace{0.1em}]},   {\bf 35}(10),  8801--8809 (2021).

\bibitem{isic}
Codella, N. C.~F., Gutman, D., Celebi, M.~E., Helba, B., Marchetti, M.~A., Dusza, S.~W., Kalloo, A., Liopyris, K., Mishra, N., Kittler, H., and Halpern, A., ``Skin lesion analysis toward melanoma detection: A challenge at the 2017 international symposium on biomedical imaging (isbi), hosted by the international skin imaging collaboration (isic),'' (2018).

\bibitem{ph2}
Mendon{\c{c}}a, T., Ferreira, P.~M., Marques, J.~S., Marcal, A.~R., and Rozeira, J., ``Ph 2-a dermoscopic image database for research and benchmarking,'' in [{\em 2013 35th annual international conference of the IEEE engineering in medicine and biology society (EMBC)}{\nolinebreak\hspace{0.1em}]},   5437--5440, IEEE (2013).

\bibitem{hemalatha2018active}
Hemalatha, R., Thamizhvani, T., Dhivya, A. J.~A., Joseph, J.~E., Babu, B., and Chandrasekaran, R., ``Active contour based segmentation techniques for medical image analysis,'' {\em Medical and Biological Image Analysis}~{\bf 4}(17),  2 (2018).

\bibitem{ravichandran2009color}
Ravichandran, K. and Ananthi, B., ``Color skin segmentation using k-means cluster,'' {\em International Journal of Computational and Applied Mathematics}~{\bf 4}(2),  153--158 (2009).

\bibitem{yogarajah2010dynamic}
Yogarajah, P., Condell, J., Curran, K., Cheddad, A., and McKevitt, P., ``A dynamic threshold approach for skin segmentation in color images,'' in [{\em 2010 IEEE International Conference on Image Processing}{\nolinebreak\hspace{0.1em}]},   2225--2228, IEEE (2010).

\bibitem{nasr2019dense}
Nasr-Esfahani, E., Rafiei, S., Jafari, M.~H., Karimi, N., Wrobel, J.~S., Samavi, S., and Soroushmehr, S.~R., ``Dense pooling layers in fully convolutional network for skin lesion segmentation,'' {\em Computerized Medical Imaging and Graphics}~{\bf 78},  101658 (2019).

\bibitem{mirikharaji2018star}
Mirikharaji, Z. and Hamarneh, G., ``Star shape prior in fully convolutional networks for skin lesion segmentation,'' in [{\em Medical Image Computing and Computer Assisted Intervention--MICCAI 2018: 21st International Conference, Granada, Spain, September 16-20, 2018, Proceedings, Part IV 11}{\nolinebreak\hspace{0.1em}]},   737--745, Springer (2018).

\bibitem{centernet}
Zhou, X., Wang, D., and Kr{\"a}henb{\"u}hl, P., ``Objects as points,'' {\em arXiv preprint arXiv:1904.07850}  (2019).

\bibitem{hagerty2019deep}
Hagerty, J.~R., Stanley, R.~J., Almubarak, H.~A., Lama, N., Kasmi, R., Guo, P., Drugge, R.~J., Rabinovitz, H.~S., Oliviero, M., and Stoecker, W.~V., ``Deep learning and handcrafted method fusion: higher diagnostic accuracy for melanoma dermoscopy images,'' {\em IEEE journal of biomedical and health informatics}~{\bf 23}(4),  1385--1391 (2019).

\bibitem{zhang2018skin}
Zhang, J., Xie, Y., Wu, Q., and Xia, Y., ``Skin lesion classification in dermoscopy images using synergic deep learning,'' in [{\em Medical Image Computing and Computer Assisted Intervention--MICCAI 2018: 21st International Conference, Granada, Spain, September 16-20, 2018, Proceedings, Part II 11}{\nolinebreak\hspace{0.1em}]},   12--20, Springer (2018).

\bibitem{mb_dcnn}
Xie, Y., Zhang, J., Xia, Y., and Shen, C., ``A mutual bootstrapping model for automated skin lesion segmentation and classification,'' {\em IEEE transactions on medical imaging}~{\bf 39}(7),  2482--2493 (2020).

\bibitem{yu2016automated}
Yu, L., Chen, H., Dou, Q., Qin, J., and Heng, P.-A., ``Automated melanoma recognition in dermoscopy images via very deep residual networks,'' {\em IEEE transactions on medical imaging}~{\bf 36}(4),  994--1004 (2016).

\bibitem{robust}
Zamir, A.~R., Sax, A., Cheerla, N., Suri, R., Cao, Z., Malik, J., and Guibas, L.~J., ``Robust learning through cross-task consistency,'' in [{\em Proceedings of the IEEE/CVF conference on computer vision and pattern recognition}{\nolinebreak\hspace{0.1em}]},   11197--11206 (2020).

\bibitem{resnet}
He, K., Zhang, X., Ren, S., and Sun, J., ``Deep residual learning for image recognition,'' in [{\em Proceedings of the IEEE conference on computer vision and pattern recognition}{\nolinebreak\hspace{0.1em}]},   770--778 (2016).

\bibitem{attention}
Vaswani, A., Shazeer, N., Parmar, N., Uszkoreit, J., Jones, L., Gomez, A.~N., Kaiser, {\L}., and Polosukhin, I., ``Attention is all you need,'' {\em Advances in neural information processing systems}~{\bf 30} (2017).

\bibitem{focal}
Lin, T.-Y., Goyal, P., Girshick, R., He, K., and Doll{\'a}r, P., ``Focal loss for dense object detection,'' in [{\em Proceedings of the IEEE international conference on computer vision}{\nolinebreak\hspace{0.1em}]},   2980--2988 (2017).

\bibitem{dice}
Sudre, C.~H., Li, W., Vercauteren, T., Ourselin, S., and Cardoso, M.~J., ``Generalised dice overlap as a deep learning loss function for highly unbalanced segmentations,'' {\em Deep Learning in Medical Image Analysis and Multimodal Learning for Clinical Decision Support} ,  240--248 (2017).

\bibitem{imagenet}
Deng, J., Dong, W., Socher, R., Li, L.-J., Li, K., and Fei-Fei, L., ``Imagenet: A large-scale hierarchical image database,'' in [{\em 2009 IEEE Conference on Computer Vision and Pattern Recognition}{\nolinebreak\hspace{0.1em}]},   248--255 (2009).

\bibitem{adam}
Kingma, D.~P. and Ba, J., ``Adam: A method for stochastic optimization,'' (2017).

\bibitem{xu2021regnet}
Xu, J., Pan, Y., Pan, X., Hoi, S., Yi, Z., and Xu, Z., ``Regnet: Self-regulated network for image classification,'' (2021).

\bibitem{effnet}
Tan, M. and Le, Q.~V., ``Efficientnet: Rethinking model scaling for convolutional neural networks,'' {\em CoRR}~{\bf abs/1905.11946} (2019).

\bibitem{svm}
Hearst, M.~A., Dumais, S.~T., Osuna, E., Platt, J., and Scholkopf, B., ``Support vector machines,'' {\em IEEE Intelligent Systems and their applications}~{\bf 13}(4),  18--28 (1998).

\bibitem{adaboost}
Schapire, R.~E., ``Explaining adaboost,'' in [{\em Empirical inference: festschrift in honor of vladimir N. Vapnik}{\nolinebreak\hspace{0.1em}]},   37--52, Springer (2013).

\bibitem{xgb}
Chen, T. and Guestrin, C., ``Xgboost: A scalable tree boosting system,'' in [{\em Proceedings of the 22nd acm sigkdd international conference on knowledge discovery and data mining}{\nolinebreak\hspace{0.1em}]},   785--794 (2016).

\end{thebibliography}
\bibliographystyle{spiebib} 
\newpage

\appendix
\section*{APPENDIX}

\section{Encoder and Decoder Architectures}\label{sec:app_enc_dec}

\begin{figure*}[b]
\centering
            \begin{subfigure}[c]{\textwidth}
                \centering
                \includegraphics[width=\textwidth]{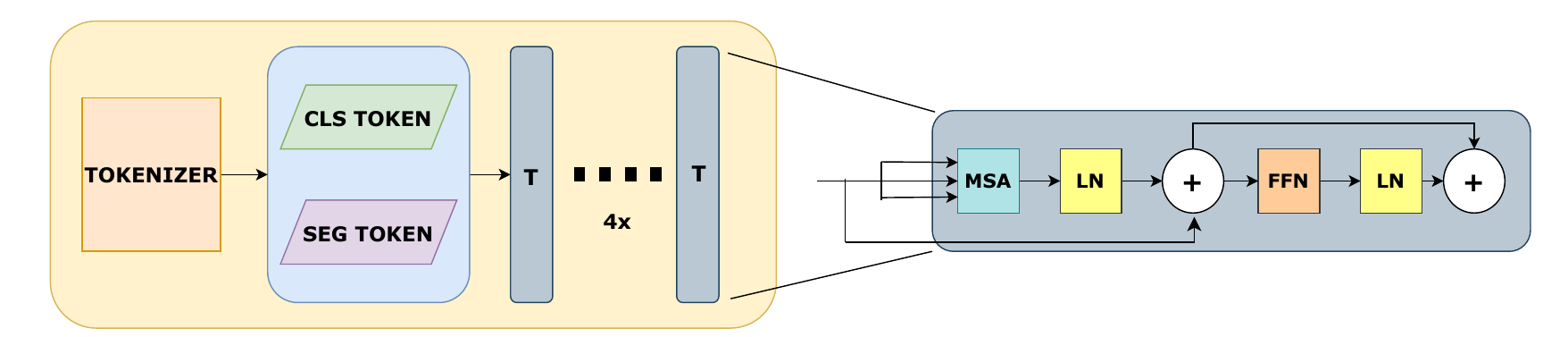}
                \caption{Vision Transformer (ViT) encoder architecture with Multi-head Self-Attention (MSA), Layer Normalization (LN) and Feed Forward Network (FFN) blocks.The MSA blocks takes in $3$ inputs: query($Q$), key($K$) and value($V$) }
                \label{fig:enc_arch}
            \end{subfigure}
        \\
        \begin{subfigure}[b]{0.49\textwidth}
            \includegraphics[width=\textwidth]{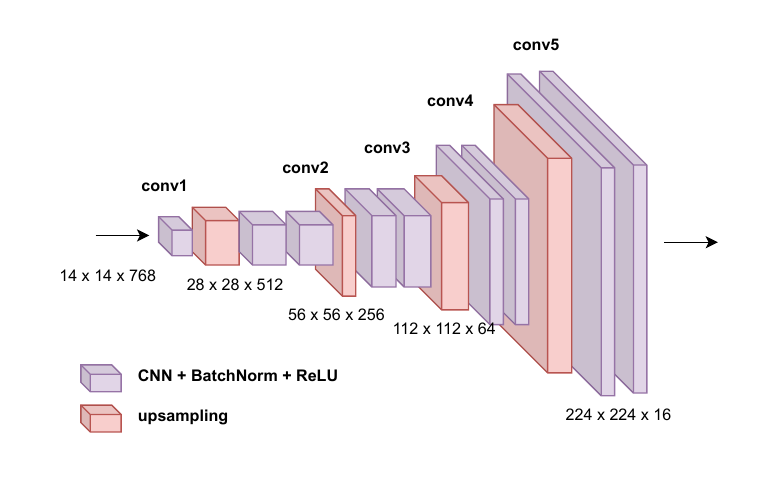}
            \caption{CNN Decoder for Segmentation Features from the segmentation tokens of transformer.}
            \label{fig:cnn_dec}
        \end{subfigure}
    \hfill
        \begin{subfigure}[b]{0.49\textwidth}
            \includegraphics[width=0.85\textwidth]{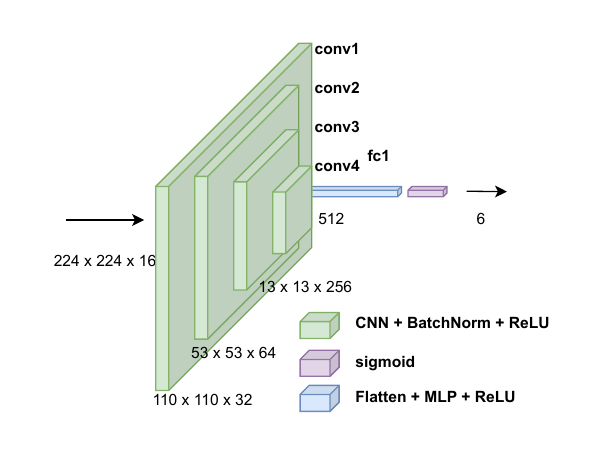}
            \caption{CNN Network for Gaussian Splat features from the segmentation features.}
            \label{fig:gauss_net}
        \end{subfigure}
    \caption{Architecture of the ViT encoder and CNN decoder for segmentation features and Gaussian Splat features.}
    \label{fig:enc_dec}
\end{figure*}

The detailed architecture for ViT encoder is given in Fig.~\ref{fig:enc_arch} where the the image is passed through the tokenizer to obtain segmentation (local) tokens and the class (global) token is a learned embedding. The tokenizer generates a token(feature vector) for a given patch of $14 \times 14$ image incase of our ViT-B/14 architecture. The tokens are concatenated along with sinusoidal position embeddings and passed through the a sequence of $4$ transformer blocks to get the feature representation for a given image as shown in Fig.~\ref{fig:model_overview}.

For obtaining the segmentation features, we employ a CNN decoder as shown in Fig.~\ref{fig:cnn_dec}, to upsample the segmentation tokens back into the input resolution of $224 \times 224$. These features are used to get the level set function through a single CNN block of stride $1$ and output channels $1$ and the Gaussian Splat features using the CNN Network shown in Fig.~\ref{fig:gauss_net} to get feature vector of size $6$ consisting of position, scale and rotation values.
\section{Analysis on Statistical Models}\label{sec:app_stat}

We perform skin cancer classification using SVM~\cite{svm}, AdaBoost~\cite{adaboost}, XGBoost~\cite{xgb} and Logistic Regression and report the top $6$ best performing models in Table~\ref{table:3}. We either directly pass the image data into the classifier or perform dimensionality reduction using Principal Component Analysis(PCA), Linear Discriminant Analysis(LDA), or pre-trained neural networks such as RegNet~\cite{xu2021regnet}, EffNet~\cite{effnet} and ViT~\cite{vit}. Using these features, the classifiers are trained on the training set and their performance is evaluated on the testset similar to other methods in the section~\ref{sec:methods}.

\renewcommand{\arraystretch}{1.5}
\begin{table*}[!h]
\centering
\resizebox{\textwidth}{!}{ 
\begin{tabular}{l|c|c|c|c|c|c|c|c|c|c|c}
\hline
\multirow{2}{*}{\textbf{Models}} & \multicolumn{5}{c|} {\textbf{Melanoma Classification}} & \multicolumn{5}{c|} {\textbf{Keratosis Classification}} & \textbf{Average} \\
\cline{2-11}
 & \textbf{M\_AC} & \textbf{M\_AUC} & \textbf{M\_AP} & \textbf{M\_SE} & \textbf{M\_SP} & \textbf{K\_AC} & \textbf{K\_AUC} & \textbf{K\_AP} & \textbf{K\_SE} & \textbf{K\_SP} & \textbf{Accuracy} \\
\Xhline{1.5pt}
SVM & 0.800 & 0.657 & 0.295 & 0.000 & 1.000 & 0.767 & 0.834 & 0.713 & 0.167 & 1.000 & 0.567 \\
\hline
AdaBoost & 0.787 & 0.589 & 0.268 & 0.000 & 0.983 & 0.773 & 0.731 & 0.592 & 0.262 & 0.972 & 0.573 \\
\hline
XGBoost + PCA & 0.820 & 0.699 & 0.381 & 0.167 & 0.983 & 0.767 & 0.866 & 0.698 & 0.333 & 0.935 & 0.613 \\
\Xhline{1.5pt}
SVM + ViT & 0.813 & 0.800 & 0.538 & 0.100 & 0.992 & 0.853 & 0.931 & 0.867 & 0.500 & 0.991 & 0.673 \\
\hline
XGBoost + EffNet & 0.820 & 0.779 & 0.518 & 0.233 & 0.967 & 0.840 & 0.898 & 0.813 & 0.500 & 0.972 & 0.687 \\
\hline
Logistic\_Reg + ViT & 0.827 & 0.809 & 0.567 & 0.367 & 0.942 & 0.840 & 0.916 & 0.832 & 0.762 & 0.870 & 0.713 \\
\Xhline{1.5pt}
\end{tabular}
} 

\center\caption{Melanoma and Keratosis Classification Results for statistical models. AC, AUC, AP, SE, and SP refer to Accuracy, Area Under Curve, Average Precision, Sensitivity, and Specificity respectively.}
\label{table:3}
\end{table*}

The performance of statistical methods are not up to the mark and they fail especially for melanoma classification as they overfit for the more frequently occurring keratosis class. Therefore, these models are neither reliable nor robust for critical skin cancer diagnosis.

\end{document}